\documentclass[11pt,a4paper]{article}
\usepackage[hyperref]{emnlp2017}
\usepackage{times}
\usepackage{latexsym}
\usepackage{hyperref}

\usepackage{amsmath, amssymb}
\usepackage{bm}
\usepackage[pdftex]{graphicx}
\usepackage{url}
\usepackage{multirow}
\usepackage{wrapfig}

\emnlpfinalcopy % Uncomment this line for the final submission

\title{A Joint Many-Task Model:\\Growing a Neural Network for Multiple NLP Tasks}

\author{Kazuma Hashimoto\thanks{~~Work was done while the first author was an intern at Salesforce Research.}, Caiming Xiong\thanks{~~Corresponding author.}, Yoshimasa Tsuruoka, and Richard Socher  \\
The University of Tokyo \\
\texttt{\{hassy, tsuruoka\}@logos.t.u-tokyo.ac.jp} \\
Salesforce Research \\
\texttt{\{cxiong, rsocher\}@salesforce.com} \\
}

\begin{document}

\maketitle

\begin{abstract}
Transfer and multi-task learning have traditionally focused on either a single source-target pair or very few, similar tasks.
Ideally, the linguistic levels of morphology, syntax and semantics would benefit each other by being trained in a single model.
We introduce a joint many-task model together with a strategy for successively growing its depth to solve increasingly complex tasks.
Higher layers include shortcut connections to lower-level task predictions to reflect linguistic hierarchies.
We use a simple regularization term to allow for optimizing all model weights to improve one task's loss without exhibiting catastrophic interference of the other tasks.
Our single end-to-end model obtains state-of-the-art or competitive results on five different tasks from tagging, parsing, relatedness, and entailment tasks.
\end{abstract}

%%%%
\section{Introduction}

\begin{figure}[t]
	\begin{center}
    	\includegraphics[width=75mm]{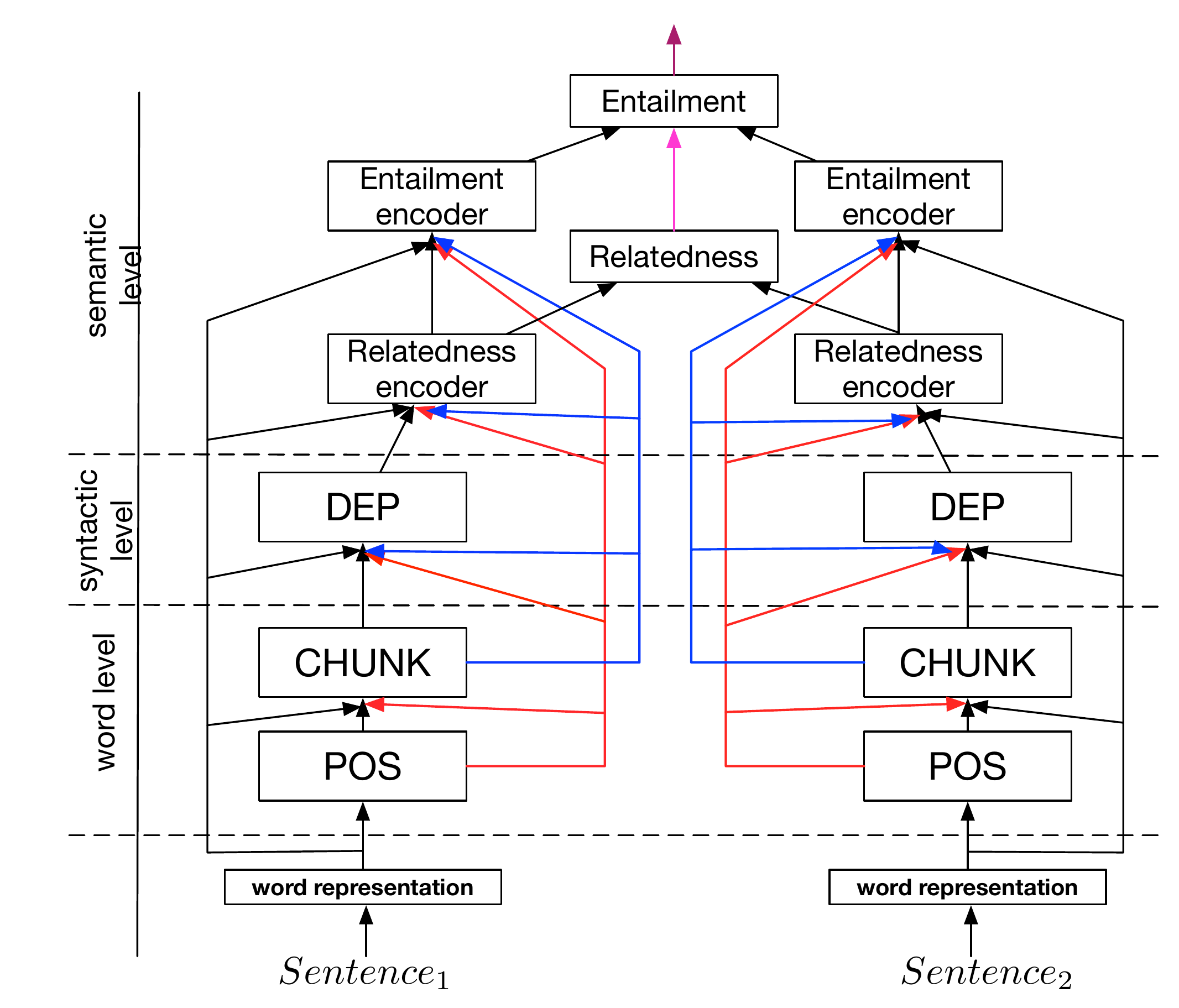}
    \end{center}
\label{fig1}

\caption{Overview of the joint many-task model predicting different linguistic outputs at successively deeper layers.}
\end{figure}

The potential for leveraging multiple levels of representation has been demonstrated in various ways in the field of Natural Language Processing (NLP).
For example, Part-Of-Speech (POS) tags are used for syntactic parsers.
The parsers are used to improve higher-level tasks, such as natural language inference~\citep{chen2016snli} and machine translation~\citep{eriguchi2016}.
These systems are often pipelines and not trained end-to-end.

Deep NLP models have yet shown benefits from predicting many increasingly complex tasks each at a successively deeper layer.
Existing models often ignore linguistic hierarchies by predicting different tasks either entirely separately or at the same depth~\citep{collobert2011senna}.

We introduce a Joint Many-Task (JMT) model, outlined in Figure~\ref{fig1}, which predicts increasingly complex NLP tasks at successively deeper layers.
Unlike traditional pipeline systems, our single JMT model can be trained end-to-end for POS tagging, chunking, dependency parsing, semantic relatedness, and textual entailment, by considering linguistic hierarchies.
We propose an adaptive training and regularization strategy to grow this model in its depth. 
With the help of this strategy we avoid catastrophic interference between the tasks.
Our model is motivated by \citet{sogaard2016} who showed that predicting two different tasks is more accurate when performed in different layers than in the same layer~\citep{collobert2011senna}.
Experimental results show that our single model achieves competitive results for all of the five different tasks, demonstrating that using linguistic hierarchies is more important than handling different tasks in the same layer.

%%%%
\section{The Joint Many-Task Model}
This section describes the inference procedure of our model, beginning at the lowest level and working our way to higher layers and more complex tasks; our model handles the five different tasks in the order of POS tagging, chunking, dependency parsing, semantic relatedness, and textual entailment, by considering linguistic hierarchies.
The POS tags are used for chunking, and the chunking tags are used for dependency parsing~\citep{Attardi2008}.
\citet{tai2015treelstm} have shown that dependencies improve the relatedness task.
The relatedness and entailment tasks are closely related to each other.
If the semantic relatedness between two sentences is very low, they are unlikely to entail each other.
Based on this observation, we make use of the information from the relatedness task for improving the entailment task.

\subsection{Word Representations}
\label{subsec:wordrep}
For each word $w_t$ in the input sentence $\bm{s}$ of length $L$, we use two types of embeddings.

\noindent
{\bf Word embeddings:}
We use Skip-gram~\citep{mikolov2013word2vec} to train word embeddings.

\noindent
{\bf Character embeddings:}
Character $n$-gram embeddings are trained by the same Skip-gram objective.
We construct the character $n$-gram vocabulary in the training data and assign an embedding for each entry.
The final character embedding is the average of the {\it unique} character $n$-gram embeddings of  $w_t$.
For example, the character $n$-grams ($n=1, 2, 3$) of the word ``Cat'' are \{C, a, t, \#B\#C, Ca, at, t\#E\#, \#B\#Ca, Cat, at\#E\#\}, where ``\#B\#'' and ``\#E\#'' represent the beginning and the end of each word, respectively.
Using the character embeddings efficiently provides morphological features.
Each word is subsequently represented as $x_t$, the concatenation of its corresponding word and character embeddings shared across the tasks.\footnote{\citet{char2017} previously proposed to train the character $n$-gram embeddings by the Skip-gram objective.}

\subsection{Word-Level Task: POS Tagging}
The first layer of the model is a bi-directional LSTM \citep{graves2005bilstm,hochreiter1997lstm} whose hidden states are used to predict POS tags.
We use the following Long Short-Term Memory (LSTM) units for the forward direction:
%\begin{equation}\label{eq:lstm}
%\begin{split}
%&i_{t}=\sigma\left(W_{i}g_{t}+b_{i}\right),~f_{t}=\sigma\left(W_{f}g_{t}+b_{f}\right),\\
%&o_{t}=\sigma\left(W_{o}g_{t}+b_{o}\right),~u_{t}=\tanh\left(W_{u}g_{t}+b_{u}\right),\\
%&c_{t}=i_{t}\odot u_{t}+f_{t}\odot c_{t-1},~h_{t}=o_{t}\odot\tanh\left(c_{t}\right),
%\end{split}
%\end{equation}
\begin{eqnarray}\label{eq:lstm}
i_{t}=\sigma\left(W_{i}g_{t}+b_{i}\right),~f_{t}=\sigma\left(W_{f}g_{t}+b_{f}\right),\nonumber\\
o_{t}=\sigma\left(W_{o}g_{t}+b_{o}\right),~u_{t}=\tanh\left(W_{u}g_{t}+b_{u}\right),\\
c_{t}=i_{t}\odot u_{t}+f_{t}\odot c_{t-1},~h_{t}=o_{t}\odot\tanh\left(c_{t}\right),\nonumber
\end{eqnarray}
where we define the input $g_t$ as $g_t=[\overrightarrow{h}_{t-1}; x_{t}]$, i.e. the concatenation of the previous hidden state and the word representation of $w_t$.
The backward pass is expanded in the same way, but a different set of weights are used.

For predicting the POS tag of $w_t$, we use the concatenation of the forward and backward states in a one-layer bi-LSTM layer corresponding to the $t$-th word: $h_{t} = [\overrightarrow{h}_{t}; \overleftarrow{h}_{t}]$.
Then each $h_{t}$ $(1\leq t \leq L)$ is fed into a standard $\mathrm{softmax}$ classifier with a single $\mathrm{ReLU}$ layer which outputs the probability vector $y^{(1)}$ for each of the POS tags.

\subsection{Word-Level Task: Chunking}

Chunking is also a word-level classification task which assigns a chunking tag ({\tt B-NP}, {\tt I-VP}, etc.) for each word. The tag specifies the region of major phrases (e.g., noun phrases) in the sentence.

Chunking is performed in the second bi-LSTM layer on top of the POS layer. 
When stacking the bi-LSTM layers, we use Eq.~(\ref{eq:lstm}) with input $g^{(2)}_t = [h^{(2)}_{t-1}; h^{(1)}_{t}; x_t; y_t^{(pos)}]$, where $h^{(1)}_{t}$ is the hidden state of the first (POS) layer.
We define the weighted label embedding $y_t^{(pos)}$ as follows:
\begin{equation}\label{eq:posEmb}
y_t^{(pos)} = \sum_{j=1}^{C}p(y^{(1)}_t=j|h_t^{(1)})\ell(j),
\end{equation}
where $C$ is the number of the POS tags, $p(y^{(1)}_t=j|h_t^{(1)})$ is the probability value that the $j$-th POS tag is assigned to $w_t$, and $\ell(j)$ is the corresponding label embedding.
The probability values are predicted by the POS layer, and thus no gold POS tags are needed.
This output embedding is similar to the $K$-best POS tag feature which has been shown to be effective in syntactic tasks~\citep{andor2016,alberti2016}.
For predicting the chunking tags, we employ the same strategy as POS tagging by using the concatenated bi-directional hidden states $h^{(2)}_{t} = [\overrightarrow{h}^{(2)}_{t}; \overleftarrow{h}^{(2)}_{t}]$ in the chunking layer. 
We also use a single $\mathrm{ReLU}$ hidden layer before the softmax classifier.

\subsection{Syntactic Task: Dependency Parsing}
\if0
\begin{figure}[t]
	\begin{center}
    	\includegraphics[width=75mm]{dep.pdf}
    \end{center}
\label{fig5}
\caption{Overview of dependency parsing in the third layer of the JMT model.}
\end{figure}
\fi
Dependency parsing identifies syntactic relations (such as an adjective modifying a noun) between word pairs in a sentence.
We use the third bi-LSTM layer to classify relations between all pairs of words.
The input vector for the LSTM includes hidden states, word representations, and the label embeddings for the two previous tasks:
$g^{(3)}_t = [h^{(3)}_{t-1}; h^{(2)}_{t}; x_t; (y_t^{(pos)} + y_t^{(chk)})]$, where we computed the chunking vector in a similar fashion as the POS vector in Eq.~(\ref{eq:posEmb}).

We predict the parent node ({\it head}) for each word.
Then a dependency label is predicted for each child-parent pair.
This approach is related to \citet{biaffine2017} and \citet{zhang2017head}, where the main difference is that our model works on a multi-task framework.
To predict the parent node of $w_t$, we define a matching function between $w_t$ and the candidates of the parent node as $m\left(t, j\right)={h^{(3)}_{t}} \cdot (W_d h^{(3)}_{j})$,
where $W_d$ is a parameter matrix.
For the root, we define $h^{(3)}_{L+1}=r$ as a parameterized vector.
To compute the probability that $w_j$ (or the root node) is the parent of $w_t$, the scores are normalized:
\begin{equation}
p(j|h^{(3)}_{t})=\frac{\exp\left(m\left(t, j\right)\right)}{\sum_{k=1, k\neq t}^{L+1}\exp\left(m\left(t, k\right)\right)}.
\end{equation}

The dependency labels are predicted using $[h^{(3)}_t; h^{(3)}_{j}]$ as input to a $\mathrm{softmax}$ classifier with a single $\mathrm{ReLU}$ layer.
We greedily select the parent node and the dependency label for each word.
When the parsing result is not a well-formed tree, we apply the first-order Eisner's algorithm~\citep{eisner1996} to obtain a well-formed tree from it.

\subsection{Semantic Task: Semantic relatedness}

\if0
\begin{figure}[t]
	\begin{center}
    	\includegraphics[width=80mm]{relatedness.pdf}
    \end{center}
\label{fig6}
\caption{Overview of the semantic tasks in the top layers of the JMT model.}
\end{figure}
\fi

The next two tasks model the semantic relationships between two input sentences. 
The first task measures the semantic relatedness between two sentences.
The output is a real-valued relatedness score for the input sentence pair.
The second task is textual entailment, which requires one to determine whether a premise sentence entails a hypothesis sentence.
There are typically three classes: entailment, contradiction, and neutral.
We use the fourth and fifth bi-LSTM layer for the relatedness  and entailment task, respectively.

Now it is required to obtain the sentence-level representation rather than the word-level representation $h^{(4)}_{t}$ used in the first three tasks.
We compute the sentence-level representation $h^{(4)}_{\mathbf{s}}$ as the element-wise maximum values across all of the word-level representations in the fourth layer:
\begin{equation}
{h}^{(4)}_{\mathbf{s}}=\max\left({h}^{(4)}_{1}, {h}^{(4)}_{2}, \ldots, {h}^{(4)}_{L}\right).
\end{equation}
This max-pooling technique has proven effective in text classification tasks~\citep{lai2015maxpooling}.

To model the semantic relatedness between $s$ and $s'$, we follow \citet{tai2015treelstm}. The feature vector for representing the semantic relatedness is computed as follows:
\begin{equation}
\label{eq:sim}
d_1(s, s')=\left[\left|{h}^{(4)}_{\mathbf{s}}-{h}^{(4)}_{\mathbf{s}'}\right|; {h}^{(4)}_{\mathbf{s}}\odot{h}^{(4)}_{\mathbf{s}'}\right],
\end{equation}
where $\left|{h}^{(4)}_{\mathbf{s}}-{h}^{(4)}_{\mathbf{s}'}\right|$ is the absolute values of the element-wise subtraction, and ${h}^{(4)}_{\mathbf{s}}\odot{h}^{(4)}_{\mathbf{s}'}$ is the element-wise multiplication.
Then $d_1(s, s')$ is fed into a $\mathrm{softmax}$ classifier with a single $\mathrm{Maxout}$ hidden layer~\citep{goodfellow2013} to output a relatedness score (from 1 to 5 in our case).

\subsection{Semantic Task: Textual entailment}
For entailment classification, we also use the max-pooling technique as in the semantic relatedness task.
To classify the premise-hypothesis pair $(s, s')$ into one of the three classes, we compute the feature vector $d_2(s, s')$ as in Eq.~(\ref{eq:sim}) except that we do not use the absolute values of the element-wise subtraction, because we need to identify which is the premise (or hypothesis).
Then $d_2(s, s')$ is fed into a $\mathrm{softmax}$ classifier.

To use the output from the relatedness layer directly, we use the label embeddings for the relatedness task.
More concretely, we compute the class label embeddings for the semantic relatedness task similar to Eq.~(\ref{eq:posEmb}).
The final feature vectors that are concatenated and fed into the entailment classifier are the weighted relatedness label embedding and the feature vector $d_2(s, s')$.
We use three $\mathrm{Maxout}$ hidden layers before the classifier.

%%%%
\section{Training the JMT Model}
\label{sec:training}
The model is trained jointly over all datasets.
During each epoch, the optimization iterates over each full training dataset in the same order as the corresponding tasks described in the modeling section.

\subsection{Pre-Training Word Representations}
\label{sec:trainingCharVecs}
We pre-train word embeddings using the Skip-gram model with negative sampling~\citep{mikolov2013word2vec}.
We also pre-train the character $n$-gram embeddings using Skip-gram.\footnote{The training code and the pre-trained embeddings are available at \url{https://github.com/hassyGo/charNgram2vec}.}
The only difference is that each input word embedding is replaced with its corresponding average character $n$-gram embedding described in Section~\ref{subsec:wordrep}.
These embeddings are fine-tuned during the model training.
We denote the embedding parameters as $\theta_{e}$.

\subsection{Training the POS Layer}
\label{subsec:trainingPOS}
Let $\theta_{\mathrm{POS}}=(W_{\mathrm{POS}}, b_{\mathrm{POS}}, \theta_{e})$ denote the set of model parameters associated with the POS layer, where $W_{\mathrm{POS}}$ is the set of the weight matrices in the first bi-LSTM and the classifier, and $b_{\mathrm{POS}}$ is the set of the bias vectors.
The objective function to optimize $\theta_{\mathrm{POS}}$ is defined as follows:
\begin{equation}
\begin{split}
J_1(\theta_{\mathrm{POS}})=&-\sum_{s}\sum_{t}\log{p(y_t^{(1)}=\alpha|h_t^{(1)})}\\
&+\lambda\|W_{\mathrm{POS}}\|^2+\delta\|\theta_{e}-\theta_{e}'\|^2,
\end{split}
\end{equation}
where $p(y_t^{(1)}=\alpha_{w_t}|h_t^{(1)})$ is the probability value that the correct label $\alpha$ is assigned to $w_t$ in the sentence $s$, $\lambda\|W_{\mathrm{POS}}\|^2$ is the L2-norm regularization term, and $\lambda$ is a hyperparameter.

We call the second regularization term $\delta\|\theta_{e}-\theta_{e}'\|^2$ a {\it successive} regularization term.
The successive regularization is based on the idea that we do not want the model to forget the information learned for the other tasks.
In the case of POS tagging, the regularization is applied to $\theta_{e}$, and $\theta_{e}'$ is the embedding parameter after training the final task in the top-most layer at the previous training epoch.
$\delta$ is a hyperparameter.

\subsection{Training the Chunking Layer}
\label{subsec:trainingChunking}
The objective function is defined as follows:
\begin{equation}
\begin{split}
J_2(\theta_{\mathrm{chk}})&=-\sum_{s}\sum_{t}\log{p(y_t^{(2)}=\alpha|h_t^{(2)})}\\
&+\lambda\|W_{\mathrm{chk}}\|^2+\delta\|\theta_{\mathrm{POS}}-\theta_{\mathrm{POS}}'\|^2,
\end{split}
\end{equation}
which is similar to that of POS tagging, and $\theta_{\mathrm{chk}}$ is $(W_{\mathrm{chk}}, b_{\mathrm{chk}}, E_{\mathrm{POS}}, \theta_{e})$, where $W_{\mathrm{chk}}$ and $b_{\mathrm{chk}}$ are the weight and bias parameters including those in $\theta_{\mathrm{POS}}$, and $E_{\mathrm{POS}}$ is the set of the POS label embeddings.
$\theta_{\mathrm{POS}}'$ is the one after training the POS layer at the current training epoch.

\subsection{Training the Dependency Layer}
The objective function is defined as follows:
\begin{equation}
\begin{split}
J_3&(\theta_{\mathrm{dep}})=-\sum_{s}\sum_{t}\log{p(\alpha|h_t^{(3)})p(\beta|h_t^{(3)},h_\alpha^{(3)})}\\
&+\lambda(\|W_{\mathrm{dep}}\|^2+\|W_d\|^2)+\delta\|\theta_{\mathrm{chk}}-\theta_{\mathrm{chk}}'\|^2,
\end{split}
\end{equation}
where $p(\alpha|h_t^{(3)})$ is the probability value assigned to the correct parent node $\alpha$ for $w_t$, and $p(\beta|h_t^{(3)}, h_\alpha^{(3)})$ is the probability value assigned to the correct dependency label $\beta$ for the child-parent pair $(w_t, \alpha)$.
$\theta_{\mathrm{dep}}$ is defined as $(W_{\mathrm{dep}}, b_{\mathrm{dep}}, W_d, r,  E_{\mathrm{POS}}, E_{\mathrm{chk}}, \theta_{e})$, where $W_{\mathrm{dep}}$ and $b_{\mathrm{dep}}$ are the weight and bias parameters including those in $\theta_{\mathrm{chk}}$, and $E_{\mathrm{chk}}$ is the set of the chunking label embeddings.

\subsection{Training the Relatedness Layer}
Following \citet{tai2015treelstm}, the objective function is defined as follows:
\begin{equation}
\begin{split}
J_4(\theta_{\mathrm{rel}})=&\sum_{(s, s')}\mathrm{KL}\left(\hat{p}(s, s')\middle\|p(h^{(4)}_{s}, h^{(4)}_{s'})\right)\\
&+\lambda\|W_{\mathrm{rel}}\|^2+\delta\|\theta_{\mathrm{dep}}-\theta_{\mathrm{dep}}'\|^2,
\end{split}
\end{equation}
where $\hat{p}(s, s')$ is the gold distribution over the defined relatedness scores, $p(h^{(4)}_{s}, h^{(4)}_{s'})$ is the predicted distribution given the the sentence representations, and $\mathrm{KL}\left(\hat{p}(s, s')\middle\|p(h^{(4)}_{s}, h^{(4)}_{s'})\right)$ is the KL-divergence between the two distributions.
$\theta_{\mathrm{rel}}$ is defined as $(W_{\mathrm{rel}}, b_{\mathrm{rel}}, E_{\mathrm{POS}}, E_{\mathrm{chk}}, \theta_{e})$.

\subsection{Training the Entailment Layer}
The objective function is defined as follows:
\begin{equation}
\begin{split}
J_5(\theta_{\mathrm{ent}})=&-\sum_{(s, s')}\log{p(y^{(5)}_{(s, s')}=\alpha|h_{s}^{(5)}, h_{s'}^{(5)})}\\
&+\lambda\|W_{\mathrm{ent}}\|^2+\delta\|\theta_{\mathrm{rel}}-\theta_{\mathrm{rel}}'\|^2,
\end{split}
\end{equation}
where $p(y^{(5)}_{(s, s')}=\alpha|h_{s}^{(5)}, h_{s'}^{(5)})$ is the probability value that the correct label $\alpha$ is assigned to the premise-hypothesis pair $(s, s')$.
$\theta_{\mathrm{ent}}$ is defined as $(W_{\mathrm{ent}}, b_{\mathrm{ent}}, E_{\mathrm{POS}}, E_{\mathrm{chk}}, E_{\mathrm{rel}}, \theta_{e})$, where $E_{\mathrm{rel}}$ is the set of the relatedness label embeddings.

%%%%
\section{Related Work}
Many deep learning approaches have proven to be effective in a variety of NLP tasks and are becoming more and more complex.
They are typically designed to handle single tasks, or some of them are designed as general-purpose models~\citep{kumar2016dmn,sutskever2014seq2seq} but applied to different tasks independently.

For handling multiple NLP tasks, multi-task learning models with deep neural networks have been proposed~\citep{collobert2011senna,luong2016mtl}, and more recently \citet{sogaard2016} have suggested that using different layers for different tasks is more effective than using the same layer in jointly learning closely-related tasks, such as POS tagging and chunking.
However, the number of tasks was limited or they have very similar task settings like word-level tagging, and it was not clear how lower-level tasks could be also improved by combining higher-level tasks.

More related to our work, \citet{godwin2016multi} also followed \citet{sogaard2016} to jointly learn POS tagging, chunking, and language modeling, and \citet{zhang2016stackprop} have shown that it is effective to jointly learn POS tagging and dependency parsing by sharing internal representations.
In the field of relation extraction, \citet{miwa2016rel} proposed a joint learning model for entity detection and relation extraction.
All of them suggest the importance of multi-task learning, and we investigate the potential of handling different types of NLP tasks rather than closely-related ones in a single hierarchical deep model.

In the field of computer vision, some transfer and multi-task learning approaches have also been proposed~\citep{li2016multi,misra2016multi}.
For example, \citet{misra2016multi} proposed a multi-task learning model to handle different tasks.
However, they assume that each data sample has annotations for the different tasks, and do not explicitly consider task hierarchies.

Recently, \citet{rusu2016progressive} have proposed a progressive neural network model to handle multiple reinforcement learning tasks, such as Atari games.
Like our JMT model, their model is also successively trained according to different tasks using different layers called columns in their paper.
In their model, once the first task is completed, the model parameters for the first task are fixed, and then the second task is handled with new model parameters.
Therefore, accuracy of the previously trained tasks is never improved.
In NLP tasks, multi-task learning has the potential to improve not only higher-level tasks, but also lower-level tasks.
Rather than fixing the pre-trained model parameters, our successive regularization allows our model to continuously train the lower-level tasks without significant accuracy drops.

%%%%
\section{Experimental Settings}

\subsection{Datasets}
\label{subsec:datasets}

\noindent
{\bf POS tagging:}
To train the POS tagging layer, we used the Wall Street Journal (WSJ) portion of Penn Treebank, and followed the standard split for the training (Section 0-18), development (Section 19-21), and test (Section 22-24) sets.
The evaluation metric is the word-level accuracy.

\noindent
{\bf Chunking:}
For chunking, we also used the WSJ corpus, and followed the standard split for the training (Section 15-18) and test (Section 20) sets as in the CoNLL 2000 shared task.
We used Section 19 as the development set and employed the IOBES tagging scheme.
The evaluation metric is the F1 score defined in the shared task.

\noindent
{\bf Dependency parsing:}
We also used the WSJ corpus for dependency parsing, and followed the standard split for the training (Section 2-21), development (Section 22), and test (Section 23) sets.
We obtained Stanford style dependencies using the version 3.3.0 of the Stanford converter.
The evaluation metrics are the Unlabeled Attachment Score (UAS) and the Labeled Attachment Score (LAS), and punctuations are excluded for the evaluation.

\noindent
{\bf Semantic relatedness:}
For the semantic relatedness task, we used the SICK dataset~\citep{marelli2014}, and followed the standard split for the training, development, and test sets.
The evaluation metric is the Mean Squared Error (MSE) between the gold and predicted scores.

\noindent
{\bf Textual entailment:}
For textual entailment, we also used the SICK dataset and exactly the same data split as the semantic relatedness dataset.
The evaluation metric is the accuracy.

\subsection{Training Details}
\label{subsec:train_detail}
We set the dimensionality of the embeddings and the hidden states in the bi-LSTMs to 100.
At each training epoch, we trained our model in the order of POS tagging, chunking, dependency parsing, semantic relatedness, and textual entailment.
We used mini-batch stochastic gradient decent and empirically found it effective to use a gradient clipping method with growing clipping values for the different tasks; concretely, we employed the simple function: $\min(3.0, depth)$, where $depth$ is the number of bi-LSTM layers involved in each task, and $3.0$ is the maximum value.
We applied our successive regularization to our model, along with L2-norm regularization and dropout~\citep{dropout2014ver}.
More details are summarized in the supplemental material.

%%%%
\section{Results and Discussion}

Table~\ref{table:main} shows our results on the test sets of the five tasks.\footnote{In chunking evaluation, we only show the results of ``Single'' and ``JMT$_{\mathrm{AB}}$'' because the sentences for chunking evaluation overlap the training data for dependency parsing.}
The column ``Single'' shows the results of handling each task separately using single-layer bi-LSTMs, and the column ``JMT$_{\mathrm{all}}$'' shows the results of our JMT model.
The single task settings only use the annotations of their own tasks.
For example, when handling dependency parsing as a single task, the POS and chunking tags are {\it not} used.
We can see that all results of the five tasks are improved in our JMT model, which shows that our JMT model can handle the five different tasks in a single model.
Our JMT model allows us to access arbitrary information learned from the different tasks.
If we want to use the model just as a POS tagger, we can use only first bi-LSTM layer.

Table~\ref{table:main} also shows the results of five subsets of the different tasks.
For example, in the case of ``JMT$_{\mathrm{ABC}}$'', only the first three layers of the bi-LSTMs are used to handle the three tasks.
In the case of ``JMT$_{\mathrm{DE}}$'', only the top two layers are used as a two-layer bi-LSTM by omitting all information from the first three layers.
The results of the closely-related tasks (``AB'', ``ABC'', and ``DE'') show that our JMT model improves both of the high-level and low-level tasks.
The results of ``JMT$_{\mathrm{CD}}$'' and ``JMT$_{\mathrm{CE}}$'' show that the parsing task can be improved by the semantic tasks.

It should be noted that in our analysis on the greedy parsing results of the ``JMT$_{\mathrm{ABC}}$'' setting, we have found that more than 95\% are well-formed dependency trees on the development set.
In the 1,700 sentences of the development data, 11 results have multiple root notes, 11 results have no root nodes, and 61 results have cycles.
These 83 parsing results are converted into well-formed trees by Eisner's algorithm, and the accuracy does not significantly change (UAS: 94.52\%$\rightarrow$94.53\%, LAS: 92.61\%$\rightarrow$92.62\%).

\begin{table*}[t]
  \begin{center}
{\small
	\begin{tabular}{cl|c|c|ccccc}
	  &             & Single   & JMT$_{\mathrm{all}}$ & JMT$_{\mathrm{AB}}$  & JMT$_{\mathrm{ABC}}$  & JMT$_{\mathrm{DE}}$ & JMT$_{\mathrm{CD}}$ & JMT$_{\mathrm{CE}}$ \\ \hline
	A $\uparrow$ & POS          &  97.45   & 97.55   & 97.52  & 97.54 & n/a & n/a & n/a \\ \hline
	B $\uparrow$ & Chunking &     95.02   &  n/a & 95.77  & n/a & n/a & n/a & n/a \\ \hline
	\multirow{2}{*}{C $\uparrow$}
	& Dependency UAS&          93.35 & 94.67  & n/a          & 94.71 & n/a & 93.53 &  93.57  \\
	& Dependency LAS&          91.42  & 92.90  & n/a         & 92.92 & n/a & 91.62 &  91.69 \\ \hline
	D $\downarrow$ & Relatedness & 0.247 &  0.233 & n/a & n/a & 0.238 & 0.251  & n/a \\ \hline
	E $\uparrow$ & Entailment  & 81.8  &  86.2  & n/a & n/a & 86.8 & n/a & 82.4 \\ \hline
  \end{tabular}
}
    \caption{Test set results for the five tasks.
    		 In the relatedness task, the lower scores are better.}
    \label{table:main}
  \end{center}
\end{table*}

\subsection{Comparison with Published Results}

\paragraph{POS tagging}
Table~\ref{table:pos} shows the results of POS tagging, and our JMT model achieves the score close to the state-of-the-art results.
The best result to date has been achieved by \citet{ling2015charlstm}, which uses character-based LSTMs.
Incorporating the character-based encoders into our JMT model would be an interesting direction, but we have shown that the simple pre-trained character $n$-gram embeddings lead to the promising result.

\begin{table*}[t!]
\begin{minipage}[t]{.32\textwidth}
  \begin{center}
{\scriptsize
	\begin{tabular}{l|c}
    Method   & Acc. $\uparrow$ \\ \hline
	JMT$_{\mathrm{all}}$ &  97.55  \\ \hline
	\citet{ling2015charlstm} &  {\bf 97.78}  \\
	\citet{kumar2016dmn} &        97.56 \\
    \citet{ma2016seqlabel} & 97.55 \\
    \citet{sogaard2011pos} & 97.50 \\
    \citet{collobert2011senna} & 97.29 \\
    \citet{tsuruoka2011} & 97.28 \\
    \citet{toutanova2003pos} & 97.27 \\ \hline
  \end{tabular}
}
    \caption{POS tagging results.}
    \label{table:pos}
  \end{center}
\end{minipage}
%
%\hspace{-5mm}
%
\begin{minipage}[t]{.32\textwidth}
  \begin{center}
{\scriptsize
	\begin{tabular}{l|c}
    Method   & F1 $\uparrow$ \\ \hline
	JMT$_{\mathrm{AB}}$ &  {\bf 95.77}  \\
   	Single				 &  95.02  \\ \hline
	\citet{sogaard2016} &  95.56  \\
	\citet{suzuki2008chunk} & 95.15 \\
    \citet{collobert2011senna} & 94.32 \\
    \citet{kudo2001chunk} & 93.91 \\
    \citet{tsuruoka2011} & 93.81 \\ \hline
  \end{tabular}
}
    \caption{Chunking results.}
    \label{table:chunk}
  \end{center}
\end{minipage}
%
%\hfill
%
\begin{minipage}[t]{.35\textwidth}
  \begin{center}
{\scriptsize
	\begin{tabular}{l|cc}
    Method   & UAS $\uparrow$ & LAS $\uparrow$ \\ \hline
	JMT$_{\mathrm{all}}$ &  94.67 & 92.90 \\
    Single & 93.35 & 91.42 \\ \hline
	\citet{biaffine2017} & {\bf 95.74} & {\bf 94.08} \\
	\citet{andor2016} &  94.61 & 92.79  \\
	\citet{alberti2016} & 94.23 & 92.36 \\
    \citet{zhang2017head} & 94.10 & 91.90 \\
    \citet{weiss2015dep} & 93.99 & 92.05 \\
    \citet{dyer2015dep} & 93.10 & 90.90 \\
    \citet{bohnet:2010dep} & 92.88 & 90.71 \\ \hline
  \end{tabular}
}
    \caption{Dependency results.}
    \label{table:dep}
  \end{center}
\end{minipage}

\end{table*}

\begin{table*}[t!]

\begin{minipage}[t]{.45\textwidth}
  \begin{center}
{\scriptsize
	\begin{tabular}{l|c}
    Method   & MSE $\downarrow$ \\ \hline
	JMT$_{\mathrm{all}}$ &  {\bf 0.233}  \\
    JMT$_{\mathrm{DE}}$ &  0.238  \\ \hline
	\citet{zhou2016coling} &  0.243  \\
    \citet{tai2015treelstm} & 0.253 \\ \hline
  \end{tabular}
}
    \caption{Semantic relatedness results.}
    \label{table:relate}
  \end{center}
\end{minipage}
\hfill
\begin{minipage}[t]{.45\textwidth}
  \begin{center}
{\scriptsize
	\begin{tabular}{l|c}
    Method   & Acc. $\uparrow$ \\ \hline
	JMT$_{\mathrm{all}}$ &  86.2  \\
	JMT$_{\mathrm{DE}}$ &  {\bf 86.8}  \\ \hline
	\citet{yin2016abcnn} &  86.2  \\
    \citet{lai2014semeval} & 84.6 \\ \hline
  \end{tabular}
}
    \caption{Textual entailment results.}
    \label{table:entail}
  \end{center}
\end{minipage}
  
\end{table*}

\paragraph{Chunking}
Table~\ref{table:chunk} shows the results of chunking, and our JMT model achieves the state-of-the-art result.
\citet{sogaard2016} proposed to jointly learn POS tagging and chunking in different layers, but they only showed improvement for chunking.
By contrast, our results show that the low-level tasks are also improved.

\paragraph{Dependency parsing}
Table~\ref{table:dep} shows the results of dependency parsing by using only the WSJ corpus in terms of the dependency annotations.\footnote{\citet{choe2016dep} employed a tri-training method to expand the training data with 400,000 trees in addition to the WSJ data, and they reported 95.9 UAS and 94.1 LAS by converting their constituency trees into dependency trees.
\citet{kuncoro2017} also reported high accuracy (95.8 UAS and 94.6 LAS) by using a converter.
}
It is notable that our simple greedy dependency parser outperforms the model in \citet{andor2016} which is based on beam search with global information.
The result suggests that the bi-LSTMs efficiently capture global information necessary for dependency parsing.
Moreover, our single task result already achieves high accuracy without the POS and chunking information.
The best result to date has been achieved by the model propsoed in \citet{biaffine2017}, which uses higher dimensional representations than ours and proposes a more sophisticated attention mechanism called {\it biaffine attention}.
It should be promising to incorporate their attention mechanism into our parsing component.

\paragraph{Semantic relatedness}
Table~\ref{table:relate} shows the results of the semantic relatedness task, and our JMT model achieves the state-of-the-art result.
The result of ``JMT$_{\mathrm{DE}}$'' is already better than the previous state-of-the-art results.
Both of \citet{zhou2016coling} and \citet{tai2015treelstm} explicitly used syntactic trees, and \citet{zhou2016coling} relied on attention mechanisms.
However, our method uses the simple max-pooling strategy, which suggests that it is worth investigating such simple methods before developing complex methods for simple tasks.
Currently, our JMT model does not explicitly use the learned dependency structures, and thus the explicit use of the output from the dependency layer should be an interesting direction of future work.

\paragraph{Textual entailment}
Table~\ref{table:entail} shows the results of textual entailment, and our JMT model achieves the state-of-the-art result.
The previous state-of-the-art result in \citet{yin2016abcnn} relied on attention mechanisms and dataset-specific data pre-processing and features.
Again, our simple max-pooling strategy achieves the state-of-the-art result boosted by the joint training.
These results show the importance of jointly handling related tasks.

\subsection{Analysis on the Model Architectures}
We investigate the effectiveness of our model in detail.
All of the results shown in this section are the development set results.

\begin{table}[t]
{\scriptsize
	\begin{center}
	\begin{tabular}{l|c|ccc}
  			   & JMT$_{\mathrm{all}}$ & w/o SC & w/o LE & w/o SC\&LE \\ \hline
    POS        & 97.88 & 97.79 		& 97.85 & 97.87\\ \hline
    Chunking   & 97.59 & 97.08 		& 97.40 & 97.33 \\ \hline
    Dependency UAS & 94.51 & 94.52 	& 94.09 & 94.04 \\
    Dependency LAS & 92.60 & 92.62 	& 92.14 & 92.03  \\ \hline
    Relatedness    & 0.236 & 0.698 	& 0.261 & 0.765 \\ \hline
    Entailment     & 84.6  & 75.0 	& 81.6  & 71.2  \\ \hline
  \end{tabular}
  \end{center}
}
    \caption{Effectiveness of the Shortcut Connections (SC) and the Label Embeddings (LE).}
    \label{tb:shortcut_comp}
\end{table}

\begin{table}[t]
{\scriptsize
	\begin{center}
  \begin{tabular}{l|c|cc}
  			  		 & JMT$_{\mathrm{ABC}}$  & w/o SC\&LE  & All-3  \\ \hline
    POS        	     & 97.90				 & 97.87       & 97.62 \\ \hline
    Chunking         & 97.80				 & 97.41       & 96.52 \\ \hline
    Dependency UAS   & 94.52				 & 94.13       & 93.59 \\
    Dependency LAS   & 92.61				 & 92.16       & 91.47 \\ \hline
  \end{tabular}
  \end{center}
}
    \caption{Effectiveness of using different layers for different tasks.}
    \label{tb:diff_arch}
\end{table}

\paragraph{Shortcut connections}
Our JMT model feeds the word representations into all of the bi-LSTM layers, which is called the shortcut connection.
Table~\ref{tb:shortcut_comp} shows the results of ``JMT$_{\mathrm{all}}$'' with and without the shortcut connections.
The results without the shortcut connections are shown in the column of ``w/o SC''.
These results clearly show that the importance of the shortcut connections, and in particular, the semantic tasks in the higher layers strongly rely on the shortcut connections.
That is, simply stacking the LSTM layers is not sufficient to handle a variety of NLP tasks in a single model.
In the supplementary material, it is qualitatively shown how the shortcut connections work in our model.

\paragraph{Output label embeddings}
Table~\ref{tb:shortcut_comp} also shows the results without using the output labels of the POS, chunking, and relatedness layers, in the column of ``w/o LE''.
These results show that the explicit use of the output information from the classifiers of the lower layers is important in our JMT model.
The results in the column of ``w/o SC\&LE'' are the ones without both of the shortcut connections and the label embeddings.

\paragraph{Different layers for different tasks}
Table~\ref{tb:diff_arch} shows the results of our ``JMT$_{\mathrm{ABC}}$'' setting and that of not using the shortcut connections and the label embeddings (``w/o SC\&LE'') as in Table~\ref{tb:shortcut_comp}.
In addition, in the column of ``All-3'', we show the results of using the highest (i.e., the third) layer for all of the three tasks without any shortcut connections and label embeddings, and thus the two settings ``w/o SC\&LE'' and ``All-3'' require exactly the same number of the model parameters.
The ``All-3'' setting is similar to the multi-task model of \citet{collobert2011senna} in that task-specific output layers are used but most of the model parameters are shared.
The results show that using the same layers for the three different tasks hampers the effectiveness of our JMT model, and the design of the model is much more important than the number of the model parameters.

\paragraph{Successive regularization}
In Table~\ref{tb:successive_reg}, the column of ``w/o SR'' shows the results of omitting the successive regularization terms described in Section~\ref{sec:training}.
We can see that the accuracy of chunking is improved by the successive regularization, while other results are not affected so much.
The chunking dataset used here is relatively small compared with other low-level tasks, POS tagging and dependency parsing.
Thus, these results suggest that the successive regularization is effective when dataset sizes are imbalanced.

\begin{table}[t]
{\scriptsize
	\begin{center}
	\begin{tabular}{l|c|c|c}
  			   & JMT$_{\mathrm{all}}$ & w/o SR & w/o VC \\ \hline
    POS        & 97.88 & 97.85		& 97.82 \\ \hline
    Chunking   & 97.59 & 97.13		& 97.45 \\ \hline
    Dependency UAS & 94.51 & 94.46 	& 94.38 \\
    Dependency LAS & 92.60 & 92.57 	& 92.48 \\ \hline
    Relatedness    & 0.236 & 0.239 	& 0.241 \\ \hline
    Entailment     & 84.6  & 84.2	& 84.8  \\ \hline
  \end{tabular}
  \end{center}
}
    \caption{Effectiveness of the Successive Regularization (SR) and the Vertical Connections (VC).}
    \label{tb:successive_reg}
\end{table}

\begin{table}[t]
{\scriptsize
	\begin{center}
	\begin{tabular}{l|c|c}
  			   & JMT$_{\mathrm{all}}$ & Random \\ \hline
    POS        & 97.88 & 97.83	\\ \hline
    Chunking   & 97.59 & 97.71	\\ \hline
    Dependency UAS & 94.51 & 94.66 \\
    Dependency LAS & 92.60 & 92.80 	\\ \hline
    Relatedness    & 0.236 & 0.298 	\\ \hline
    Entailment     & 84.6  & 83.2	\\ \hline
  \end{tabular}
  \end{center}
}
    \caption{Effects of the order of training.}
    \label{tb:random}
\end{table}

\begin{table}[t]
{\scriptsize
	\begin{center}
	\begin{tabular}{l|cc}
  			   & Single & Single+ \\ \hline
    POS        & \multicolumn{2}{|c}{97.52} \\ \hline
    Chunking   & 95.65 & 96.08 \\ \hline
    Dependency UAS & 93.38 & 93.88 \\
    Dependency LAS & 91.37 & 91.83 \\ \hline
    Relatedness    & 0.239 & 0.665 \\ \hline
    Entailment     & 83.8  & 66.4 \\ \hline
  \end{tabular}
  \end{center}
}
    \caption{Effects of depth for the {\it single} tasks.}
    \label{tb:single_plus}
\end{table}

\begin{table}[t]
{\scriptsize
	\begin{center}
	\begin{tabular}{l|c|c}
 	Single 	   & W\&C & Only W \\ \hline
    POS        & 97.52 & 96.26 \\ \hline
    Chunking   & 95.65 & 94.92 \\ \hline
    Dependency UAS & 93.38 & 92.90 \\
    Dependency LAS & 91.37 & 90.44 \\ \hline
  \end{tabular}
  \end{center}
}
    \caption{Effects of the character embeddings.}
    \label{tb:char_comp}
\end{table}

\paragraph{Vertical connections}
We investigated our JMT results without using the vertical connections in the five-layer bi-LSTMs.
More concretely, when constructing the input vectors $g_t$, we do not use the bi-LSTM hidden states of the previous layers.
Table~\ref{tb:successive_reg} also shows the JMT$_{\mathrm{all}}$ results with and without the vertical connections.
As shown in the column of ``w/o VC'', we observed the competitive results.
Therefore, in the target tasks used in our model, sharing the word representations and the output label embeddings is more effective than just stacking the bi-LSTM layers.

\paragraph{Order of training}
Our JMT model iterates the training process in the order described in Section~\ref{sec:training}.
Our hypothesis is that it is important to start from the lower-level tasks and gradually move to the higher-level tasks.
Table~\ref{tb:random} shows the results of training our model by randomly shuffling the order of the tasks for each epoch in the column of ``Random''.
We see that the scores of the semantic tasks drop by the random strategy.
In our preliminary experiments, we have found that constructing the mini-batch samples from different tasks also hampers the effectiveness of our model, which also supports our hypothesis.

\paragraph{Depth}
The single task settings shown in Table~\ref{table:main} are obtained by using single layer bi-LSTMs, but in our JMT model, the higher-level tasks use successively deeper layers.
To investigate the gap between the different number of the layers for each task, we also show the results of using multi-layer bi-LSTMs for the single task settings, in the column of ``Single+'' in Table~\ref{tb:single_plus}.
More concretely, we use the same number of the layers with our JMT model; for example, three layers are used for dependency parsing, and five layers are used for textual entailment.
As shown in these results, deeper layers do not always lead to better results, and the joint learning is more important than making the models complex only for single tasks.

\paragraph{Character $n$-gram embeddings}
Finally, Table~\ref{tb:char_comp} shows the results for the three single tasks with and without the pre-trained character $n$-gram embeddings.
The column of ``W\&C'' corresponds to using both of the word and character $n$-gram embeddings, and that of ``Only W'' corresponds to using only the word embeddings.
These results clearly show that jointly using the pre-trained word and character $n$-gram embeddings is helpful in improving the results.
The pre-training of the character $n$-gram embeddings is also effective;
for example, without the pre-training, the POS accuracy drops from 97.52\% to 97.38\% and the chunking accuracy drops from 95.65\% to 95.14\%.

\subsection{Discussion}

\paragraph{Training strategies}
In our JMT model, it is not obvious when to stop the training while trying to maximize the scores of all the five tasks.
We focused on maximizing the accuracy of dependency parsing on the development data in our experiments.
However, the sizes of the training data are different across the different tasks; for example, the semantic tasks include only 4,500 sentence pairs, and the dependency parsing dataset includes 39,832 sentences with word-level annotations.
Thus, in general, dependency parsing requires more training epochs than the semantic tasks, but currently, our model trains all of the tasks for the same training epochs.
The same strategy for decreasing the learning rate is also shared across all the different tasks, although our growing gradient clipping method described in Section~\ref{subsec:train_detail} helps improve the results.
Indeed, we observed that better scores of the semantic tasks can be achieved before the accuracy of dependency parsing reaches the best score.
Developing a method for achieving the best scores for all of the tasks at the same time is important future work.

\paragraph{More tasks}
Our JMT model has the potential of handling more tasks than the five tasks used in our experiments; examples include entity detection and relation extraction as in \citet{miwa2016rel} as well as language modeling~\citep{godwin2016multi}.
It is also a promising direction to train each task for multiple domains by focusing on domain adaptation~\citep{sogaard2016}.
In particular, incorporating language modeling tasks provides an opportunity to use large text data.
Such large text data was used in our experiments to pre-train the word and character $n$-gram embeddings. However, it would be preferable to efficiently use it for improving the entire model.

\paragraph{Task-oriented learning of low-level tasks}
Each task in our JMT model is supervised by its corresponding dataset.
However, it would be possible to learn low-level tasks by optimizing high-level tasks, because the model parameters of the low-level tasks can be directly modified by learning the high-level tasks.
One example has already been presented in \citet{hashimoto2017lgp}, where our JMT model is extended to learning task-oriented latent graph structures of sentences by training our dependency parsing component according to a neural machine translation objective.

%%%%
\section{Conclusion}
We presented a joint many-task model to handle multiple NLP tasks with growing depth in a single end-to-end model.
Our model is successively trained by considering linguistic hierarchies, directly feeding word representations into all layers, explicitly using low-level predictions, and applying successive regularization.
In experiments on five NLP tasks, our single model achieves the state-of-the-art or competitive results on chunking, dependency parsing, semantic relatedness, and textual entailment.

\subsubsection*{Acknowledgments}
We thank the anonymous reviewers and the Salesforce Research team members for their fruitful comments and discussions.

\bibliography{allBibsFinal}
\bibliographystyle{emnlp_natbib.bst}

\appendix

%%%%
\section*{Supplemental Material}

%%%%
\section{Training Details}

\paragraph{Pre-training embeddings}
We used the {\tt word2vec} toolkit to pre-train the word embeddings.
We created our training corpus by selecting lowercased English Wikipedia text and obtained 100-dimensional Skip-gram word embeddings trained with the context window size 1, the negative sampling method (15 negative samples), and the sub-sampling method ($10^{-5}$ of the sub-sampling coefficient).
We also pre-trained the character $n$-gram embeddings using the same parameter settings with the case-sensitive Wikipedia text.
We trained the character $n$-gram embeddings for $n=1, 2, 3, 4$ in the pre-training step.

\paragraph{Embedding initialization}
We used the pre-trained word embeddings to initialize the word embeddings, and the word vocabulary was built based on the training data of the five tasks.
All words in the training data were included in the word vocabulary, and we employed the {\it word-dropout} method~\citep{kiperwasser2016} to train the word embedding for the unknown words.
We also built the character $n$-gram vocabulary for $n=2, 3, 4$, following \citet{wieting2016}, and the character $n$-gram embeddings were initialized with the pre-trained embeddings.
All of the label embeddings were initialized with uniform random values in $[-\sqrt{6/(dim+C)}, \sqrt{6/(dim+C)}]$, where $dim=100$ is the dimensionality of the label embeddings and $C$ is the number of labels.

\paragraph{Weight initialization}
The dimensionality of the hidden layers in the bi-LSTMs was set to 100.
We initialized all of the softmax parameters and bias vectors, except for the forget biases in the LSTMs, with zeros, and the weight matrix $W_d$ and the root node vector $r$ for dependency parsing were also initialized with zeros.
All of the forget biases were initialized with ones.
The other weight matrices were initialized with uniform random values in $[-\sqrt{6/(row+col)}, \sqrt{6/(row+col)}]$, where $row$ and $col$ are the number of rows and columns of the matrices, respectively.

\paragraph{Optimization}
At each epoch, we trained our model in the order of POS tagging, chunking, dependency parsing, semantic relatedness, and textual entailment.
We used mini-batch stochastic gradient decent to train our model.
The mini-batch size was set to 25 for POS tagging, chunking, and the SICK tasks, and 15 for dependency parsing.
We used a gradient clipping strategy with growing clipping values for the different tasks; concretely, we employed the simple function: $\min(3.0, depth)$, where $depth$ is the number of bi-LSTM layers involved in each task, and $3.0$ is the maximum value.
The learning rate at the $k$-th epoch was set to $\frac{\varepsilon}{1.0+\rho (k-1)}$, where $\varepsilon$ is the initial learning rate, and $\rho$ is the hyperparameter to decrease the learning rate.
We set $\varepsilon$ to 1.0 and $\rho$ to 0.3.
At each epoch, the same learning rate was shared across all of the tasks.

\paragraph{Regularization}
We set the regularization coefficient to $10^{-6}$ for the LSTM weight matrices, $10^{-5}$ for the weight matrices in the classifiers, and $10^{-3}$ for the successive regularization term excluding the classifier parameters of the lower-level tasks, respectively.
The successive regularization coefficient for the classifier parameters was set to $10^{-2}$.
We also used {\it dropout}~\citep{dropout2014ver}.
The dropout rate was set to 0.2 for the vertical connections in the multi-layer bi-LSTMs~\citep{pham2015dropout}, the word representations and the label embeddings of the entailment layer, and the classifier of the POS tagging, chunking, dependency parsing, and entailment.
A different dropout rate of 0.4 was used for the word representations and the label embeddings of the POS, chunking, and dependency layers, and the classifier of the relatedness layer.

%%%%
\section{Details of Character $N$-Gram Embeddings}
\label{sec:char_detail}

Here we first describe the pre-training process of the character $n$-gram embeddings in detail and then show further analysis on the results in Table~12.

\subsection{Pre-Training with Skip-Gram Objective}
We pre-train the character $n$-gram embeddings using the objective function of the Skip-gram model with negative sampling~\citep{mikolov2013word2vec}.
We build the vocabulary of the character $n$-grams based on the training corpus, the case-sensitive English Wikipedia text.
This is because such case-sensitive information is important in handling some types of words like named entities.
Assuming that a word $w$ has its corresponding $K$ character $n$-grams $\{cn_1, cn_2, \ldots, cn_K\}$, where any overlaps and unknown ones are removed.
Then the word $w$ is represented with an embedding $v_c(w)$ computed as follows:
\begin{equation}
v_c(w)=\frac{1}{K}\sum_{i=1}^{K}v(cn_i),
\end{equation}
where $v(cn_i)$ is the parameterized embedding of the character $n$-gram $cn_i$, and the computation of $v_c(w)$ is exactly the same as the one used in our JMT model explained in Section~2.1.

The remaining part of the pre-training process is the same as the original Skip-gram model.
For each word-context pair $(w, \overline{w})$ in the training corpus, $N$ negative context words are sampled, and the objective function is defined as follows:
\begin{equation}
\begin{split}
\sum_{(w, \overline{w})}\biggl(&-\log{\sigma(v_c(w)\cdot \tilde{v}(\overline{w}))}\\
&-\sum_{i=1}^{N}\log{\sigma(-v_c(w)\cdot \tilde{v}(\overline{w}_i))}\biggr),
\end{split}
\end{equation}
where $\sigma(\cdot)$ is the logistic sigmoid function, $\tilde{v}(\overline{w})$ is the weight vector for the context word $\overline{w}$, and $\overline{w}_i$ is a negative sample.
It should be noted that the weight vectors for the context words are parameterized for the words without any character information.

\subsection{Effectiveness on Unknown Words}
\begin{table*}[t]
  \begin{center}
%{\small
	\begin{tabular}{l|c|c}
  	Single (POS) & Overall Acc. & Acc. for unknown words \\ \hline
    W\&C         & 97.52 & 90.68 (3,502/3,862) \\ \hline
    Only W       & 96.26 & 71.44 (2,759/3,862) \\ \hline
  \end{tabular}
%}
    \caption{POS tagging scores on the development set with and without the character $n$-gram embeddings, focusing on accuracy for unknown words.
    	     The overall accuracy scores are taken from Table~12.
             There are 3,862 unknown words in the sentences of the development set.}
    \label{tb:pos_unk}
  \end{center}
\end{table*}

One expectation from the use of the character $n$-gram embeddings is to better handle unknown words.
We verified this assumption in the single task setting for POS tagging, based on the results reported in Table~12.
Table~\ref{tb:pos_unk} shows that the joint use of the word and character $n$-gram embeddings improves the score by about 19\% in terms of the accuracy for unknown words.

We also show the results of the single task setting for dependency parsing in Table~\ref{tb:dep_unk}.
Again, we can see that using the character-level information is effective, and in particular, the improvement of the LAS score is large.
These results suggest that it is better to use not only the word embeddings, but also the character $n$-gram embeddings by default.
Recently, the joint use of word and character information has proven to be effective in language modeling~\citep{miyamoto2016char}, but just using the simple character $n$-gram embeddings is fast and also effective.

\begin{table*}[t]
  \begin{center}
%{\small
	\begin{tabular}{l|cc|cc}
                        & \multicolumn{2}{|c|}{Overall scores} & \multicolumn{2}{|c}{Scores for unknown words} \\
  	Single (Dependency) & UAS   & LAS   & UAS             & LAS \\ \hline
    W\&C                & 93.38 & 91.37 & 92.21 (900/976) & 87.81 (857/976) \\ \hline
    Only W              & 92.90 & 90.44 & 91.39 (892/976) & 81.05 (791/976) \\ \hline
  \end{tabular}
%}
    \caption{Dependency parsing scores on the development set with and without the character $n$-gram embeddings, focusing on UAS and LAS for unknown words.
    	     The overall scores are taken from Table~12.
             There are 976 unknown words in the sentences of the development set.}
    \label{tb:dep_unk}
  \end{center}
\end{table*}

%%%%
\section{Analysis on Dependency Parsing}
\label{sec:dep_analysis}

Our dependency parser is based on the idea of predicting a head (or parent) for each word, and thus the parsing results do not always lead to correct trees.
To inspect this aspect, we checked the parsing results on the development set (1,700 sentences), using the ``JMT$_{\mathrm{ABC}}$'' setting.

In the dependency annotations used in this work, each sentence has only one root node, and we have found 11 sentences with multiple root nodes and 11 sentences with no root nodes in our parsing results.
We show two examples below:
\begin{itemize}
\item[(a)] Underneath the headline `` Diversification , '' it \underline{{\bf counsels}} , `` Based on the events of the past week , all investors {\bf need} to know their portfolios are balanced to help protect them against the market 's volatility . ''
\item[(b)] Mr. Eskandarian , who resigned his Della Femina post in September , becomes \underline{chairman} and chief executive of Arnold .
\end{itemize}
In the example (a), the two boldfaced words ``counsels'' and ``need'' are predicted as child nodes of the root node, and the underlined word ``counsels'' is the correct one based on the gold annotations.
This example sentence (a) consists of multiple internal sentences, and our parser misunderstood that both of the two verbs are the heads of the sentence.

In the example (b), none of the words is connected to the root node, and the correct child node of the root is the underlined word ``chairman''.
Without the internal phrase ``who resigned... in September'', the example sentence (b) is very simple, but we have found that such a simplified sentence is still not parsed correctly.
In many cases, verbs are linked to the root nodes, but sometimes other types of words like nouns can be the candidates.
In our model, the single parameterized vector $r$ is used to represent the root node for each sentence.
Therefore, the results of the examples (a) and (b) suggest that it would be needed to capture various types of root nodes, and using sentence-dependent root representations would lead to better results in future work.

%%%%
\section{Analysis on Semantic Tasks}
\label{sec:sem_analysis}

We inspected the development set results on the semantic tasks using the ``JMT$_{\mathrm{all}}$'' setting.
In our model, the highest-level task is the textual entailment task.
We show an example premise-hypothesis pair which is misclassified in our results:
\begin{itemize}
\item[] Premise: ``A surfer is riding a {\it big} wave across dark green water'', and
\item[] Hypothesis: ``The surfer is riding a {\it small} wave''.
\end{itemize}
The predicted label is {\tt entailment}, but the gold label is {\tt contradiction}.
This example is very easy by focusing on the difference between the two words ``big'' and ``small''.
However, our model fails to correctly classify this example because there are few opportunities to learn the difference.
Our model relies on the pre-trained word embeddings based on word co-occurrence statistics~\citep{mikolov2013word2vec}, and it is widely known that such co-occurrence-based embeddings can rarely discriminate between antonyms and synonyms~\citep{ono2015ant}.
Moreover, the other four tasks in our JMT model do not explicitly provide the opportunities to learn such semantic aspects.
Even in the training data of the textual entailment task, we can find only one example to learn the difference between the two words, which is not enough to obtain generalization capacities.
Therefore, it is worth investigating the explicit use of external dictionaries or the use of pre-trained word embeddings learned with such dictionaries~\citep{ono2015ant}, to see whether our JMT model is further improved not only for the semantic tasks, but also for the low-level tasks.

%%%%
\section{How Do Shared Embeddings Change}
\label{sec:change}

In our JMT model, the word and character $n$-gram embedding matrices are shared across all of the five different tasks.
To better qualitatively explain the importance of the shortcut connections shown in Table~7, we inspected how the shared embeddings change when fed into the different bi-LSTM layers.
More concretely, we checked closest neighbors in terms of the cosine similarity for the word representations before and after fed into the forward LSTM layers.
In particular, we used the corresponding part of $W_u$ in Eq.~(1) to perform linear transformation of the input embeddings, because $u_t$ directly affects the hidden states of the LSTMs.
Thus, this is a context-independent analysis.

Table~\ref{tb:knn} shows the examples of the word ``standing''.
The row of ``Embedding'' shows the cases of using the shared embeddings, and the others show the results of using the linear-transformed embeddings.
In the column of ``Only word'', the results of using only the word embeddings are shown.
The closest neighbors in the case of ``Embedding'' capture the semantic similarity, but after fed into the POS layer, the semantic similarity is almost washed out.
This is not surprising because it is sufficient to cluster the words of the same POS tags: here, {\tt NN}, {\tt VBG}, etc.
In the chunking layer, the similarity in terms of verbs is captured, and this is because it is sufficient to identify the coarse chunking tags: here, {\tt VP}.
In the dependency layer, the closest neighbors are adverbs, gerunds of verbs, and nouns, and all of them can be child nodes of verbs in dependency trees.
However, this information is not sufficient in further classifying the dependency labels.
Then we can see that in the column of ``Word and char'', jointly using the character $n$-gram embeddings adds the morphological information, and as shown in Table~12, the LAS score is substantially improved.

In the case of semantic tasks, the projected embeddings capture not only syntactic, but also semantic similarities.
These results show that different tasks need different aspects of the word similarities, and our JMT model efficiently transforms the shared embeddings for the different tasks by the simple linear transformation.
Therefore, without the shortcut connections, the information about the word representations are fed into the semantic tasks after transformed in the lower layers where the semantic similarities are not always important.
Indeed, the results of the semantic tasks are very poor without the shortcut connections.

\begin{table}[t]
  \begin{center}
%{\small
	\begin{tabular}{l|l|l}
 			   & Word and char & Only word \\ \hline
    		   & leaning  & stood \\
        	   & kneeling & stands \\
    Embedding  & saluting & sit \\
    		   & clinging & pillar \\
    		   & railing  & cross-legged \\ \hline

			   & warning  & ladder \\
               & waxing   & rc6280 \\
    POS        & dunking  & bethle \\
    		   & proving  & warning \\
               & tipping  & f-a-18 \\ \hline

			   & applauding  & fight \\
               & disdaining  & favor \\
    Chunking   & pickin      & pick \\
    		   & readjusting & rejoin \\
               & reclaiming  & answer \\ \hline

			   & guaranteeing & patiently \\
               & resting      & hugging \\
	Dependency & grounding    & anxiously \\
               & hanging      & resting \\
               & hugging      & disappointment \\ \hline

							& stood           & stood \\
                            & stands          & unchallenged \\
	Relatedness			    & unchallenged    & stands \\
    						& notwithstanding & beside \\
                            & judging         & exists \\ \hline

							& nudging    & beside \\
                            & skirting   & stands \\
	Entailment			    & straddling & pillar \\
    						& contesting & swung \\
                            & footing    & ovation \\ \hline
  \end{tabular}
%}
    \caption{Closest neighbors of the word ``standing'' in the embedding space and the projected space in each forward LSTM.}
    \label{tb:knn}
  \end{center}
\end{table}

\end{document}